\documentclass[letterpaper, 10 pt, conference]{ieeeconf}  % Comment this line out if you need a4paper

\IEEEoverridecommandlockouts                              % This command is only needed if 
\usepackage{graphics} % for pdf, bitmapped graphics files
\usepackage{epsfig} % for postscript graphics files
\usepackage{mathptmx} % assumes new font selection scheme installed
\usepackage{times} % assumes new font selection scheme installed
\usepackage{mathrsfs}
\usepackage{amsmath} % assumes amsmath package installed
\usepackage{amssymb}  % assumes amsmath package installed
\usepackage{algorithm}
\usepackage{algorithmic}
\usepackage{amsfonts}       % blackboard math symbols

\usepackage{hyperref}       % hyperlinks
\usepackage{url}            % simple URL typesetting
\usepackage{xcolor}         % colors
\usepackage{tikz}
\usepackage{multirow}

\title{\LARGE \bf
Occ-LLM: Enhancing Autonomous Driving with Occupancy-Based Large Language Models
}

\author{Tianshuo Xu$^{1}$, Hao Lu$^{1}$, Xu Yan$^{2}$, Yingjie Cai$^{2}$, Bingbing Liu$^{2}$ and Yingcong Chen$^{1*}$% <-this % stops a space
\thanks{$^{1}$Tianshuo Xu, Hao Lu, and Yingcong Chen are with the department of AI Thrust, Information Hub of Hong Kong University of Science and Technology (Guangzhou). {\tt\footnotesize txu647@connect.hkust-gz.edu.cn, hlu585@connect.hkust-gz.edu.cn, yingcongchen@ust.hk}}%
\thanks{$^{2}$ Xu Yan, Yingjie Cai, and Bingbing Liu are with the Huawei Noah’s Ark Lab.
{\tt\footnotesize yanxu44@huawei.com, caiyingjie@link.cuhk.edu.hk liu.bingbing@huawei.com}}%
\thanks{*The corresponding author.}% <-this % stops a space
}

\begin{document}

\maketitle
\thispagestyle{empty}
\pagestyle{empty}

%%%%%%%%%%%%%%%%%%%%%%%%%%%%%%%%%%%%%%%%%%%%%%%%%%%%%%%%%%%%%%%%%%%%%%%%%%%%%%%%
\begin{abstract}

Large Language Models (LLMs) have made substantial advancements in the field of robotic and autonomous driving. This study presents the first Occupancy-based Large Language Model (Occ-LLM), which represents a pioneering effort to integrate LLMs with an important representation. To effectively encode occupancy as input for the LLM and address the category imbalances associated with occupancy, we propose Motion Separation Variational Autoencoder (MS-VAE). This innovative approach utilizes prior knowledge to distinguish dynamic objects from static scenes before inputting them into a tailored Variational Autoencoder (VAE). This separation enhances the model's capacity to concentrate on dynamic trajectories while effectively reconstructing static scenes. The efficacy of Occ-LLM has been validated across key tasks, including 4D occupancy forecasting, self-ego planning, and occupancy-based scene question answering. Comprehensive evaluations demonstrate that Occ-LLM significantly surpasses existing state-of-the-art methodologies, achieving gains of about 6\% in Intersection over Union (IoU) and 4\% in mean Intersection over Union (mIoU) for the task of 4D occupancy forecasting. These findings highlight the transformative potential of Occ-LLM in reshaping current paradigms within robotic and autonomous driving.

\end{abstract}

%%%%%%%%%%%%%%%%%%%%%%%%%%%%%%%%%%%%%%%%%%%%%%%%%%%%%%%%%%%%%%%%%%%%%%%%%%%%%%%%
\section{INTRODUCTION}

Large Language Models (LLMs) have evolved rapidly\cite{achiam2023gpt, touvron2023llama, team2023gemini, zeng2021pangu}, becoming integral to advancing artificial intelligence across various industries \cite{qin2023gpt_general, rao2023gpt_med, irons2023gpt_science, drapal2023gpt_law}. 
Initially designed for natural language processing, LLMs have demonstrated remarkable adaptability in complex domains such as autonomous driving due to their robust generalization capabilities \cite{chen2023driving, hu2023gaia, dewangan2023talk2bev, sima2023drivelm}. These capabilities are particularly essential for robotic or autonomous driving systems, which currently lack generalization \cite{chen2023challenge1, chib2023recent}.
Currently, LLM applications in autonomous driving mainly use image-based inputs \cite{jiang2023vad}, which lack the spatial perception needed for comprehensive environmental understanding.
Existing methods in vision-based \cite{sima2023drivelm, hu2023gaia} and LiDAR-based \cite{tang2023thma, shubodh2024lip} approaches, while enhancing vehicle navigation and environmental understanding, are computationally intensive and often lack transparency in intermediate reasoning processes.

Occupancy serves as a highly expressive modality in autonomous driving \cite{mescheder2019occupancy}, offering rich spatial and semantic insights by comprehensively representing both the foreground and background of a scene. This universal representation facilitates the perception of objects regardless of their specific categories, whether known or unidentified. Notably, leading automotive manufacturers, such as Tesla \cite{TeslaAutopilot}, are progressively adopting occupancy-based systems within their vehicles, highlighting a shift towards this robust method of environmental interpretation.

\begin{figure}[t]
    \centering
    \includegraphics[width=0.48\textwidth]{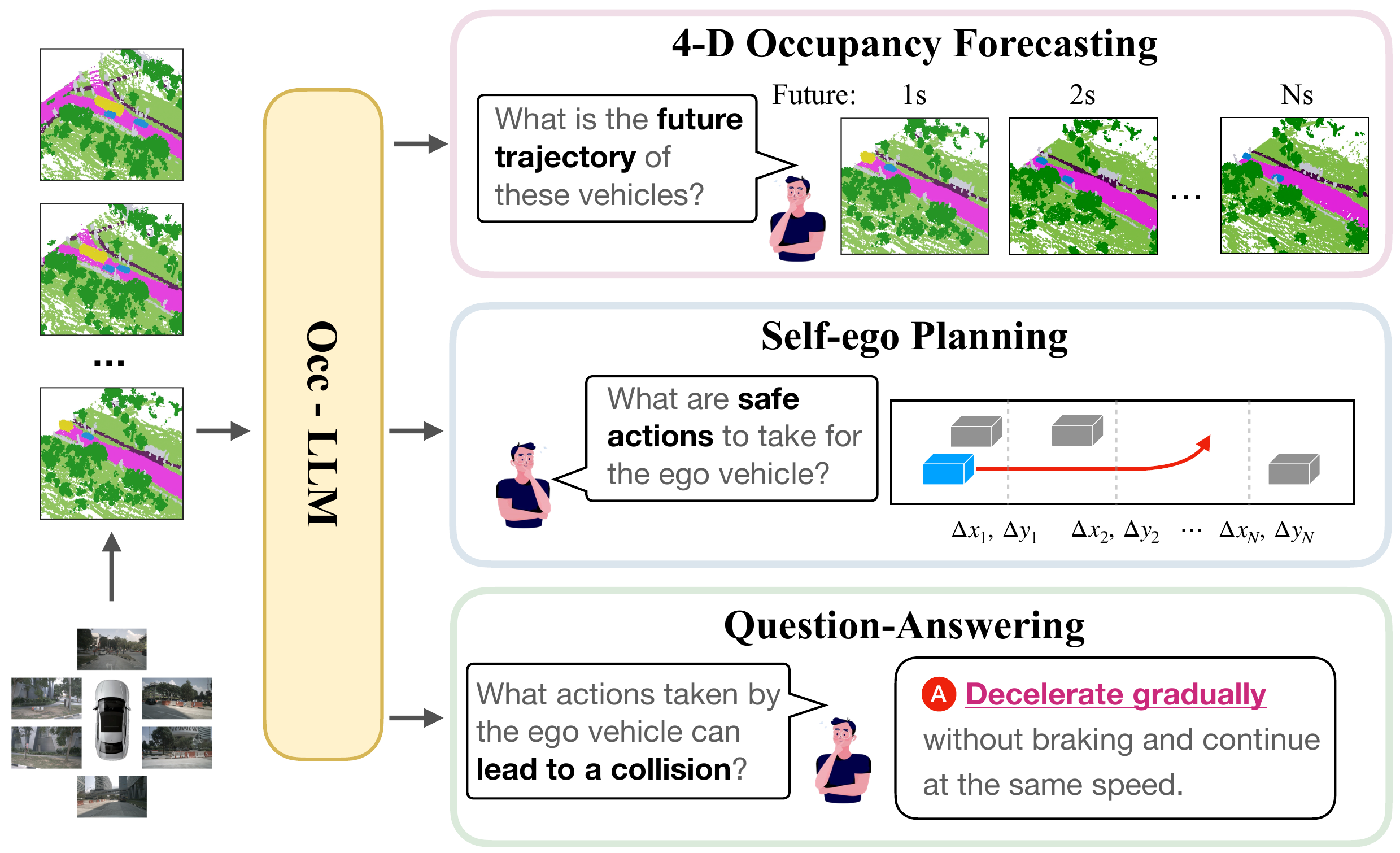}
    \caption{We present \textbf{Occ-LLM}, an occupancy-based large language model designed for autonomous driving scene prediction, planning, and understanding \textbf{(zoom in for the best view)}. 
    }
    \label{fig:teaser}
\end{figure}

This paper aims to develop a foundational model for various downstream tasks in autonomous driving by leveraging the sophisticated analytical and generalization capabilities of LLMs to interpret and utilize occupancy grids. However, direct integration of occupancy representation into LLMs is challenging due to the unbalanced occupancy categories and the predominance of voxels representing air, leading to inefficient learning and memory issues. To overcome these challenges, we propose a novel method termed the Motion Separation Variational Autoencoder (MS-VAE). This approach separates voxels associated with movable entities (e.g., cars, pedestrians) from those related to immovable structures (e.g., streets, greenery) within the occupancy scene. By doing so, it enhances the model's focus on dynamic object trajectories and improves the reconstruction of static scenes, akin to residual learning. This separation significantly reduces learning difficulties and improves overall model performance.

% Our occupancy-based large language model (Occ-LLM) is meticulously crafted to cater to a broad spectrum of applications within the domain of autonomous driving, highlighting its utility and significance in this dynamically evolving field. Principal applications of our model include 4D occupancy scene forecasting, self-ego planning, and occupancy-based scene question answering (QA). The capacity to predict occupancy across four dimensions—space and time—extends to forecasting the trajectories of other vehicles, environmental changes, and pedestrian behaviors. This predictive ability enables autonomous vehicles to execute proactive and informed decision-making. Self-ego planning leverages these forecasts to devise and implement maneuvers safely, thus ensuring robust navigational autonomy. Moreover, the occupancy-based QA functionality equips the system to adeptly interpret and adapt to dynamic changes within the driving environment, thereby enhancing real-time decision-making capabilities. These applications are integral to augmenting the safety, efficiency, and reliability of autonomous driving systems. They provide a comprehensive understanding of the vehicle’s surroundings and proactively identify potential hazards, significantly enhancing operational safety and performance. The main contributions are listed below:

Our occupancy-based large language model (Occ-LLM) is meticulously designed to cater to a diverse range of applications within the domain of autonomous driving. Principal applications of our model include 4D occupancy scene forecasting, self-ego planning, and occupancy-based scene question answering (QA), as shown in Fig.~\ref{fig:teaser}. 
These applications are integral to augmenting the safety, efficiency, and reliability of autonomous driving systems. To validate the effectiveness of our models, we conducted extensive evaluations comparing Occ-LLM to other state-of-the-art methods. Our model demonstrated superior performance, achieving a 32.52\% IoU and a 20.99\% mIoU in 4-D occupancy scene forecasting, significantly outperforming the state-of-the-art model, which achieved an IoU of 26.63\% and an mIoU of 17.14\% (the average over 3-second). For self-ego planning, our model reduced the 3-second average L2 distance to 0.28 meters, compared to the 1.17 meters achieved by the leading alternative. Additionally, in occupancy-based scene QA, Occ-LLM consistently provided accurate and reliable responses, thereby enhancing the decision-making capabilities in autonomous driving system.

The main contributions of this paper are listed below:

\begin{itemize}

    \item We introduce an occupancy-based large language model (Occ-LLM) for autonomous driving, demonstrating superior scene comprehension.

    \item We propose the Motion Separation Variational Autoencoder (MS-VAE), which manages large volumes of occupancy grid data by distinguishing between movable and immovable elements, enhancing system performance across various indicators

    \item We showcase the versatility of Occ-LLM through its applications in 4D occupancy scene forecasting, self-ego planning, and occupancy-based scene question answering, illustrating its superiority across multiple dimensions of autonomous driving. 

    \item We showcase the generalization capabilities of Occ-LLM by accessing existing occupancy prediction methods, illustrating its practicability for autonomous driving. 
    
    % and robustness of 
    
    % % a novel integration technique combining occupancy representation with LLMs through the Motion Separation Variational Autoencoder (MS-VAE). This approach efficiently manages large volumes of occupancy grid data by distinguishing between movable and immovable elements, thereby enhancing system performance across various indicators.
    
    % % operational efficiency, and adaptability under varied conditions. Our method significantly outperforms existing approaches in benchmark tests, achieving a higher IoU and mIoU by notable margins.
    
    % \item We showcase the versatility and robustness of Occ-LLM through its applications in 4D occupancy scene forecasting, self-ego planning, and occupancy-based scene question answering, illustrating its superiority across multiple dimensions of autonomous driving. 
    
    % \item Note that, our model is orthogonal to any scene-to-occupancy prediction methods, enabling seamless integration with any predicted occupancy representation.

\end{itemize}

\section{RELATED WORK}

\subsection{Multimodal Large Language Model}

Recent advancements in Multimodal Large Language Models (MLLMs) have sparked interest by combining the advanced reasoning capabilities of LLMs with image, video, and audio data~\cite{li2023blip,zhu2023minigpt,liu2023visual,lu2024gpt}. These models have shown remarkable proficiency in tasks such as zero-shot and few-shot image classification, segmentation, and object detection by leveraging the synergy between visual and textual data. In the context of autonomous driving, LLMs address a critical gap by enhancing scene understanding, providing richer semantic context, and facilitating decision-making processes, which current systems lack.
Several methods have been proposed to leverage LLMs in autonomous driving. Vision-based approaches, such as DriveGPT4, interpret video inputs to generate driving-related textual responses~\cite{chen2023driving}, while models like HiLM-D enhance hazard identification and intention prediction through high-resolution visual data~\cite{ding2023hilm}. Lidar-based methods utilize vectorized visual embeddings to equip LLMs with environmental perception capabilities, enabling detailed analysis of the driving scene~\cite{fu2023driving}.

% Despite these advancements, existing methods often require significant computational resources and lack explainability in intermediate reasoning steps. For example, Talk2BEV combines Bird's Eye View (BEV) maps with linguistic context to enable visual-linguistic reasoning in autonomous vehicles~\cite{dewangan2023talk2bev}, yet the complexity of integrating multiple modalities presents challenges in real-time applications. Moreover, generative models like GAIA-1 highlight the potential of LLMs to anticipate various outcomes based on vehicle maneuvers but underscore the computational demands of such sophisticated models~\cite{hu2023gaia}.

\subsection{Occupancy}

% The majority of prevalent 3D perception methods, whether leveraging LiDAR sweeps, multi-view images, or multi-modal data, construct BEV (Bird's Eye View) feature representations before performing various downstream tasks in the BEV space \cite{bevdet, li2022bevdepth, li2022bevformer, liu2022bevfusion, liang2022bevfusion,lu2023towards,lu2024scaling}. 
Recently, 3D semantic occupancy provides a more detailed representation of the environment by explicitly modeling the occupancy status of each voxel within a 3D grid. SSCNet \cite{song2017semantic} was the first to introduce the task of semantic scene completion, integrating geometric and semantic information. Subsequent works commonly utilize geometric inputs with explicit depth information \cite{lmscnet, aicnet, js3cnet, sketch}. MonoScene \cite{cao2022monoscene} proposed the first monocular approach for semantic scene completion, using a 3D UNet \cite{ronneberger2015u} to process voxel features generated through sight projection. Various networks based on the transfer architecture have been designed~\cite{huang2023tri,huang2023tri,Zhang_2023_ICCV}. Additionally, several concurrent works have focused on proposing surrounding-view benchmarks for 3D semantic occupancy prediction, contributing to the rapid advancement of the occupancy community~\cite{wang2023openoccupancy,wang2023openoccupancy, wei2023surroundocc, tong2023scene, tian2023occ3d}. OccWorld learns a world model based on 3D occupancy, which has attracted much attention with its interpretability and efficiency. Further, this paper attempts to use the large language model as a bridge to unify occupancy tasks.

% TPVFormer \cite{huang2023tri} introduced a tri-perspective view representation for describing 3D scenes in semantic occupancy prediction. VoxFormer \cite{li2023voxformer} presented a two-stage transformer-based semantic scene completion framework capable of producing complete 3D volumetric semantics from only 2D images. OccFormer \cite{Zhang_2023_ICCV} introduced a dual-path transformer network to effectively process 3D volumes in semantic occupancy prediction, achieving long-range, dynamic, and efficient encoding of camera-generated 3D voxel features. 

% bevdet fb-bev bevformer-occ

\section{METHODS}

\begin{figure*}[t]
    \centering
    \includegraphics[width=0.8\textwidth]{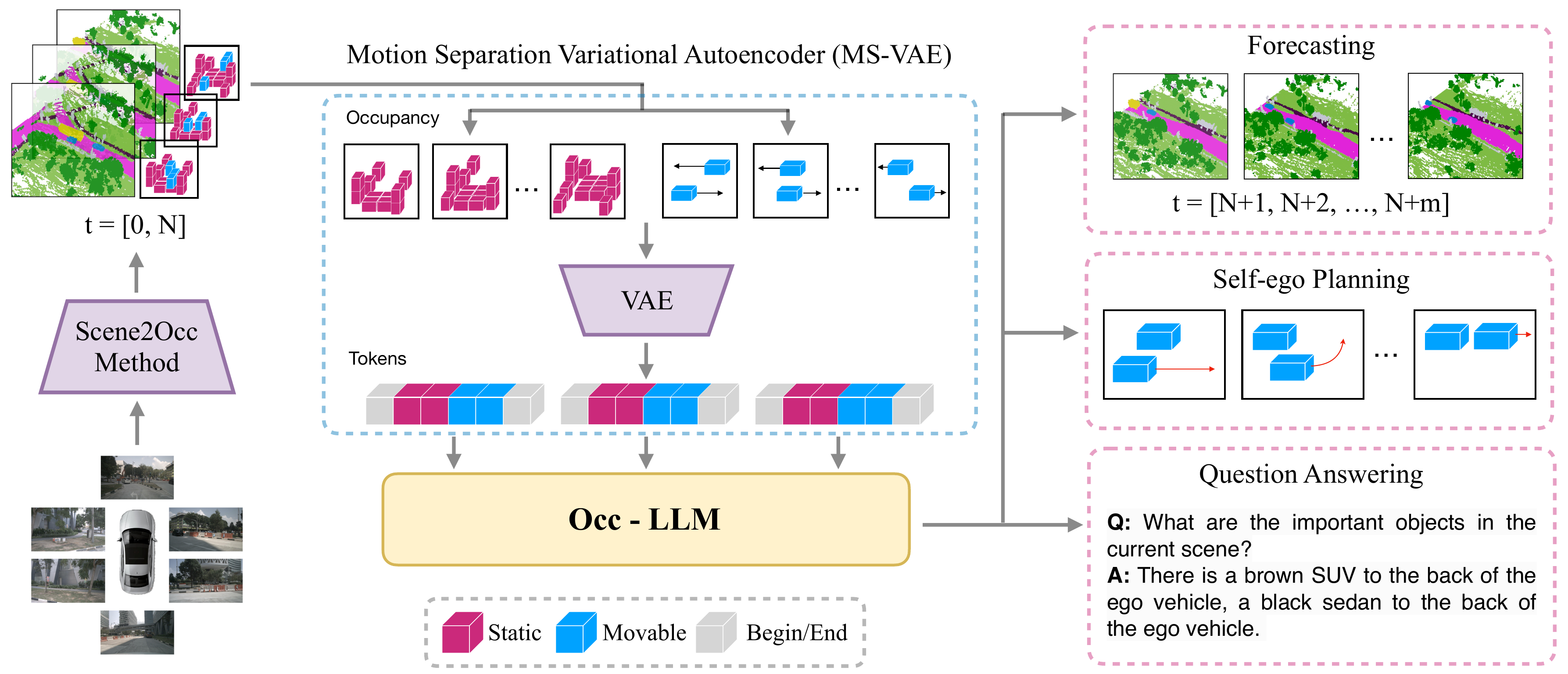}
    \caption{Overview of the proposed Occ-LLM framework. Initially, results from multiview cameras are converted into occupancy representations utilizing existing occupancy prediction algorithms. Subsequently, the Motion Separation strategy is employed to differentiate voxels associated with moving objects from static elements. These differentiated voxels are then independently encoded into latent representations using our custom-designed VAE. Finally, these latents are processed as specified in Section~\ref{sec:occ-pre} before being integrated into the LLM, completing the preparatory steps for downstream applications.}
    \label{fig:overview}
\end{figure*}

% This section introduces the Occ-LLM framework, integrating Large Language Models (LLMs) with occupancy representation to enhance autonomous driving systems (Fig.~\ref{fig:overview}). The framework uses occupancy representation to provide a detailed spatial and semantic understanding of the driving environment, improving scene interpretation and decision-making.

% We begin by converting the original multiview images into occupancy representation using existing occupancy prediction methods. In Sec.~\ref{sec:occ-vae}, we introduce the core component of our approach, the Motion Separation Variational Autoencoder (MS-VAE), which processes high-dimensional occupancy representation by distinguishing between dynamic and static elements within the scene. This differentiation reduces computational load and enhances the model's learning efficiency. We further process the occupancy latent encoded by MS-VAE in Sec.~\ref{sec:occ-pre}, ultimately flattening them for input into the LLM.

% Designed for a range of applications in autonomous driving, the Occ-LLM framework supports functionalities such as 4D occupancy forecasting, self-ego planning, and occupancy-based scene question answering, as detailed in Sec.~\ref{sec:downstream}. These capabilities enable the model to predict and respond to future changes, significantly enhancing the safety and effectiveness of autonomous vehicles.
This section introduces the Occ-LLM framework, which integrates Large Language Models (LLMs) with occupancy representation to improve autonomous driving systems (Fig.~\ref{fig:overview}). The framework enhances spatial and semantic understanding, aiding scene interpretation and decision-making. We first convert multiview images into occupancy representation using existing methods. In Sec.~\ref{sec:occ-vae}, we present the core Motion Separation Variational Autoencoder (MS-VAE), which differentiates between dynamic and static elements, reducing computational load and improving learning efficiency. The MS-VAE output is further processed and flattened for input into the LLM (Sec.~\ref{sec:occ-pre}). Designed for various autonomous driving tasks, the Occ-LLM supports 4D occupancy forecasting, self-ego planning, and occupancy-based scene question answering (Sec.~\ref{sec:downstream}), enhancing safety and effectiveness.

% This section introduces the Occ-LLM framework, which integrates Large Language Models (LLMs) with occupancy representation to improve autonomous driving systems (shown in Fig.~\ref{fig:overview}). The framework leverages occupancy as a key modality, offering a detailed spatial and semantic understanding of the driving environment to enhance scene interpretation and decision-making.

% Designed for diverse applications in autonomous driving, in Sec.~\ref{sec:downstream}, the Occ-LLM framework supports functionalities like 4D occupancy forecasting, self-ego planning, and occupancy-based scene question answering. These capabilities enable the model to predict and react to future environmental changes, significantly boosting the safety and efficacy of autonomous vehicles.

% In this section, we will discuss the implementation details and training methodology of our proposed occupancy-based large language model (Occ-LLM). 
% In this section, we propose several methods to enhance the integration of occupancy representation into large language models (LLMs), a challenging task due to the high voxel count and spatial complexity of occupancy representation. Specifically, we introduce a novel Variational Autoencoder (VAE) model that differentiates between motion voxels and immobility voxels to improve both occupancy reconstruction and generalization, as detailed in Sec.\ref{sec:occ-pre}. Additionally, we present various data augmentation techniques in Sec.~\ref{sec:occ-dataaug} to further enhance the model's generalization capabilities.

\subsection{Motion Separation Variational Autoencoder}\label{sec:occ-vae}

% Building on established multi-model LLM integration methods \cite{liu2023visual, hu2023gaia, shi2019variational}, we aim to train a Variational Autoencoder (VAE) to facilitate modal fusion and reduce the computational cost associated with occupancy representation. 
% However, directly integrating occupancy representation into LLMs presents significant challenges, primarily due to the highly unbalanced nature of occupancy categories and the predominance of voxels representing air. This imbalance results in sparse and inefficient data representations, which are not conducive to the dense and continuous data processing typical in LLMs.
% To address these challenges, we propose the Motion Separation Variational Autoencoder (MS-VAE). MS-VAE innovatively separates dynamic (movable) and static (immovable) components within the occupancy grid, allowing for targeted processing of significantly more informative regions of the environment. This separation not only enhances the efficiency of data encoding into the LLM but also improves the model’s focus on dynamic elements crucial for autonomous navigation tasks. By redefining the occupancy representation handling, MS-VAE facilitates a more balanced and effective integration into LLM frameworks.

Building on established multi-modal LLM integration methods \cite{liu2023visual, hu2023gaia, shi2019variational}, we aim to train a Variational Autoencoder (VAE) to facilitate modal fusion and reduce computational costs. Direct integration of occupancy representation into LLMs faces challenges due to unbalanced occupancy categories and the predominance of air voxels, resulting in sparse and inefficient data representations. To overcome this, we propose the Motion Separation Variational Autoencoder (MS-VAE), which separates dynamic and static components within the occupancy grid. This enhances encoding efficiency and shifts focus to dynamic elements essential for autonomous navigation. MS-VAE thus enables more balanced and effective integration into LLM frameworks.

The core concept of the Motion Separation Variational Autoencoder (MS-VAE) involves training two distinct VQ-VAEs to encode and decode moving and static occupancy voxels separately. However, we discovered that maintaining a single encoder and decoder while utilizing two different codebooks for moving and static voxels can also yield satisfactory results. For clarity, we describe this approach mathematically.

Let \( \mathbf{x} \) represent the input occupancy representation, with \( \mathbf{x}_m \) and \( \mathbf{x}_s \) denoting the moving and static voxels, respectively. The encoder \( q_\phi(z|\mathbf{x}) \) maps the input \( \mathbf{x} \) to a latent space \( \mathbf{z} \). For the MS-VAE, we define two separate latent variables \( \mathbf{z}_m \) and \( \mathbf{z}_s \) for moving and static voxels: 

\begin{equation} \mathbf{z}_m \sim q_\phi(\mathbf{z}_m | \mathbf{x}_m),\ \mathbf{z}_s \sim q_\phi(\mathbf{z}_s | \mathbf{x}_s). \end{equation}

Each encoded latent \( \mathbf{z}_m \) and \( \mathbf{z}_s \) searches in the corresponding codebook \( \mathbf{C}_m \) and \( \mathbf{C}_s \), and is replaced by the most similar codebook entry before being input to the decoder. This process is represented as:

\begin{equation} \mathbf{z}'_m = \text{argmin}_{\mathbf{c}_m \in \mathbf{C}_m} \| \mathbf{z}_m - \mathbf{c}_m \|,\ \mathbf{z}'_s = \text{argmin}_{\mathbf{c}_s \in \mathbf{C}_s} \| \mathbf{z}_s - \mathbf{c}_s \|. \end{equation}

The decoder \( p_\theta(\mathbf{x} | \mathbf{z}) \) reconstructs the input from the quantized latent variables \( \mathbf{z}'_m \) and \( \mathbf{z}'_s \):

\begin{equation} \hat{\mathbf{x}}_m = p_\theta(\mathbf{x}_m | \mathbf{z}'_m),\ \hat{\mathbf{x}}_s = p_\theta(\mathbf{x}_s | \mathbf{z}'_s). \end{equation}

To facilitate the separation of motion and static elements within the occupancy representation, we apply transformations based on the classification of voxels. Let $\mathcal{M}$ denote the set of movable classes. We define indicator functions for motion and air-filling in the modified occupancy representation as follows:

Define an indicator function $\mathbf{1}_\mathcal{M}(\mathbf{x})$ such that:
\begin{equation}
\mathbf{1}_\mathcal{M}(\mathbf{x}) = 
\begin{cases} 
1 & \text{if } \mathbf{x} \in \mathcal{M}, \\
0 & \text{otherwise}.
\end{cases}
\end{equation}

The modified motion occupancy $\mathbf{x}_m'$ and static occupancy $\mathbf{x}_s'$ are then given by:
\begin{equation}
\mathbf{x}_m' = (1 - \mathbf{1}_\mathcal{M}(\mathbf{x})) \cdot \mathbf{x}_m,
\end{equation}
\begin{equation}
\mathbf{x}_s' = \mathbf{1}_\mathcal{M}(\mathbf{x}) \cdot \text{air} + (1 - \mathbf{1}_\mathcal{M}(\mathbf{x})) \cdot \mathbf{x}_s,
\end{equation}
where $\text{air}$ denotes the representation of air in the static occupancy grid, typically encoded as a placeholder value that represents unoccupied space.

To reconstruct the raw occupancy representation, we utilize a $\text{mask} = (\hat{\mathbf{x}}_m \neq 0)$ to differentiate active motion regions. The reconstructed occupancy $\hat{\mathbf{x}}$ combines the static and motion components as follows:
\begin{equation}
\hat{\mathbf{x}} = \hat{\mathbf{x}}_m \cdot \text{mask} + \hat{\mathbf{x}}_s \cdot (1 - \text{mask}).
\end{equation}

% To facilitate the separation, we modify the occupancy representation such that motion voxels are split and the space is filled with 0 in the new motion occupancy. Conversely, the static occupancy is adjusted by filling the space (where the motion voxels were split) with 17, which denotes air. This process can be described as:

% \begin{equation}
%     \mathbf{x}_m' = \begin{cases} 
%     \mathbf{x}_m & \text{if voxel belongs to a moving object} \\
%     0 & \text{otherwise} 
%     \end{cases},
% \end{equation}

% \begin{equation}
%     \mathbf{x}_s' = \begin{cases} 
%     17 & \text{if voxel belongs to a moving object} \\
%     \mathbf{x}_s & \text{otherwise} 
%     \end{cases}.
% \end{equation}

% In the final module, to reconstruct the raw occupancy, we establish a mask where $\text{mask} = (\hat{\mathbf{x}}_m \neq 0)$. The reconstructed occupancy \( \hat{\mathbf{x}} \) is then obtained by combining the static occupancy and the masked motion occupancy:

% \begin{equation}
%     \hat{\mathbf{x}}[\text{mask}] = \hat{\mathbf{x}}_m[\text{mask}],\ \hat{\mathbf{x}}[1-\text{mask}] = \hat{\mathbf{x}}_s[1-\text{mask}].
% \end{equation}

The overall loss function for training the MS-VAE combines the reconstruction loss and the commitment loss to ensure the encoded latent is close to the codebook entries:

\begin{equation} 
\begin{split}
\mathcal{L} = \mathbb{E}_{q_\phi(\mathbf{z}_m | \mathbf{x}_m)} & \left[ \log p_\theta(\mathbf{x}_m | \mathbf{z}'_m) \right] + \mathbb{E}_{q_\phi(\mathbf{z}_s | \mathbf{x}_s)} \left[ \log p_\theta(\mathbf{x}_s | \mathbf{z}'_s) \right] \\
 + \beta & \left( \| \mathbf{z}_m - \mathbf{z}'_m \|^2 + \| \mathbf{z}_s - \mathbf{z}'_s \|^2 \right).
\end{split}
\end{equation} 

By leveraging separate codebooks for the moving and static voxels while keeping a unified encoder and decoder, and by appropriately handling the occupancy representation, the MS-VAE effectively captures the distinct characteristics of each voxel type, resulting in improved occupancy reconstruction and generalization.

In addition, the overall VAE architecture referred to the methodology outlined in OccWorld's implementation \cite{zheng2023occworld}, specifically treating the occupancy as 2D data with 16 channels and employing a 2D VAE for encoding and decoding. However, to preserve the integrity of three-dimensional information, we integrate a layer of lightweight 3D convolution both prior to the Encoder and after the Decoder. This modification respects the spatial dimensions inherent to the occupancy representation and substantially enhances the reconstructed occupancy's quality. This approach, in contrast to the conventional usage of a 2D VAE, significantly improves the fidelity of the occupancy representation in three-dimensional space.

% In addition, in the overall VAE model architecture, we refer to OccWorld's \cite{zheng2023occworld} implementation, \emph{i.e.,} treating the occupancy as a 2D data with channel=16 and training a 2D VAE to encode and decode it, but in order to minimize the loss of 3D information, we add a layer of light's 3D convolution before the Encoder and after the Decoder. Compared with simply using a 2D VAE to encode and decode the occupancy, this approach can substantially improve the reconstruction quality of the occupancy.

\subsection{Pre-processing of Integrating Occupancy with LLM}\label{sec:occ-pre}

\textbf{Patchify.} Following the encoding of raw occupancy representation using the MS-VAE, the resulting latent representation remains substantial. To address this, we adopt an approach akin to the Vision Transformer (ViT) \cite{vit} by partitioning the occupancy latent space into small grids and flattening it. Our observations indicate that the patch size significantly impacts the quality of occupancy reconstruction. This is because predicting future occupancy frames encompasses aspects of perception and low-level vision tasks. For instance, perception tasks typically benefit from larger patch sizes, facilitating a better understanding of the semantic information of the input data \cite{vit}. Conversely, low-level vision tasks often employ smaller patch sizes to achieve higher-quality data reconstruction \cite{uvit}. Through an ablation study, we determined that a patch size of 10 yields optimal results.

\begin{figure}[t]
    \centering
    \includegraphics[width=0.45\textwidth]{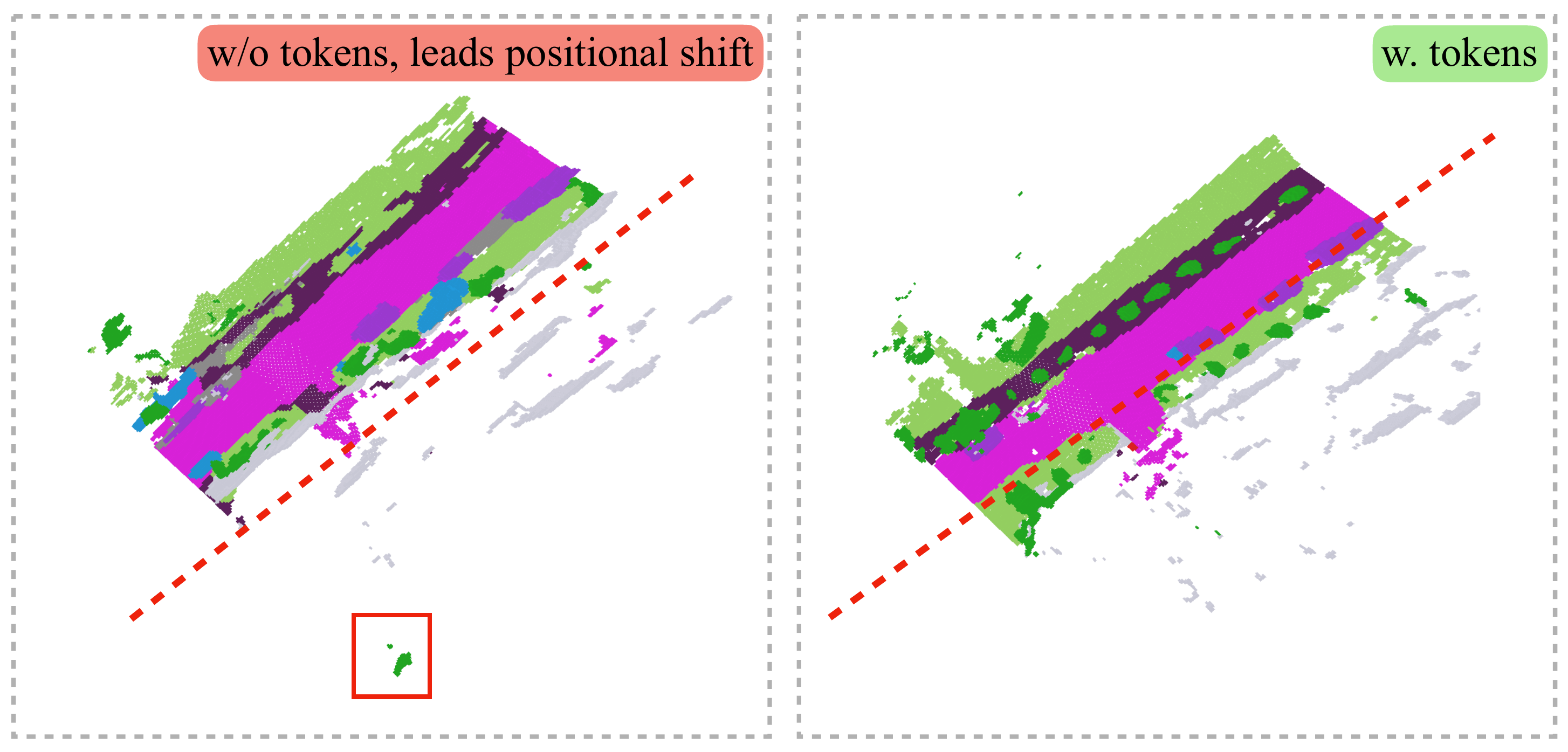}
    \caption{Illustration of the positional shift problem in occupancy representation, where the \textcolor{red}{red} dotted line represents the central axis, and the \textcolor{red}{red} box signifies the occurrence of an object in the subsequent frame that appears within the current frame. This problem is mitigated by appending tokens to the beginning \texttt{<}occ\texttt{>} and end \texttt{<}/occ\texttt{>} of each frame’s latent occupancy representation.}
    \label{fig:frame_sep}
\end{figure}

\textbf{Frame separation.} We found that the flattened occupancy latent for each frame is relatively long, and directly concatenating the flattened occupancy latent of multiple frames leads to positional drift in the generated occupancy. This drift manifests as portions of occupancy from one frame appearing in subsequent frames, causing a cascading misalignment (shown in Fig.~\ref{fig:frame_sep}).

To address this issue, we propose a straightforward but effective solution: adding specific text tokens at the beginning and end of each occupancy latent frame. Specifically, we use ``\texttt{<}occ\texttt{>}" at the beginning and ``\texttt{<}/occ\texttt{>}" at the end. These tokens delineate the intervals between frames during inference, effectively eliminating the drift problem.

\textbf{Pre-fusion.} We introduce a pre-fusion method to better establish the connection between occupancy representation and self-ego actions. This method involves first encoding self-ego actions through multiple MLP layers. Similar to the approach SE-Net \cite{senet}, we then use the encoded action latent as a weight to modulate the occupancy tents. This technique enhances the consistency between the occupancy representation and the self-ego actions, improving overall model performance.

\subsection{Downstream Tasks}\label{sec:downstream}
The Occ-LLM framework supports a variety of downstream tasks critical for enhancing autonomous driving systems, including 4D occupancy forecasting, self-ego planning, and occupancy-based scene question answering. Task switching is managed through specific prompts: ``\texttt{<}4-D occupancy forecasting and self-ego planning\texttt{>}" initiates the combined task of 4D occupancy forecasting and self-ego planning, while ``\texttt{<}question-answering\texttt{>}" triggers the question-answering task. These tasks collectively enhance situational awareness and decision-making. 4-D occupancy forecasting predicts environmental dynamics, which is crucial for anticipating hazards. Self-ego planning uses these forecasts for safe, efficient navigation. Occupancy-based scene question answering interprets complex situations, aiding in informed decision-making. Together, these capabilities significantly improve autonomous driving systems' safety, reliability, and efficiency.

\section{EXPERIMENTS}

In this section, we present an extensive set of experiments to evaluate the performance of our proposed Occ-LLM. We utilize Llama2 \cite{touvron2023llama} as the foundational model. 
We evaluate 4D occupancy forecasting using Intersection over Union (IoU) \cite{everingham2010pascal_iou} and mean Intersection over Union (mIoU) \cite{cheng2021multiscale_miou} metrics. Self-ego planning capability is assessed using the L2 distance metric.
% The evaluation of 4D occupancy forecasting is conducted using Intersection over Union (IoU) \cite{everingham2010pascal_iou} and mean Intersection over Union (mIoU) \cite{cheng2021multiscale_miou} metrics, while the self-ego planning capability is assessed using the L2 distance metric.

We employ the Nuscenes dataset \cite{caesar2020nuscenes}, which comprises 1000 scenes. These scenes are divided into 700 for training, 150 for validation, and 150 for testing. Each scene contains approximately 50 frames, corresponding to an occupancy scene. The occupancy representation has dimensions of (200,200,16), where the first (200,200) represents the length and width, and 16 represents the height. This dataset configuration enables a comprehensive assessment and validation of our model's performance across various scenarios.

\subsection{Comparisons with the State-of-the-art Methods}\label{sec:exp_occ_ego}

\begin{table*}[!t]
\centering
\caption{Quantitative results of 4D occupancy forecasting and motion planning. ``Vanilla" refers to the direct flattening of occupancy representation and its injection into the LLM for training.}
\label{tab:quanti}
\scalebox{0.9}{
\begin{tabular}{llcccccccccccc}
\hline
\multirow{2}{*}{Methods} & \multirow{2}{*}{Input} & \multicolumn{4}{c}{IoU $\uparrow$ (\%)} & \multicolumn{4}{c}{mIoU $\uparrow$ (\%)} & \multicolumn{4}{c}{L2 $\downarrow$ (m)} \\ \cline{3-14} 
 &  & 1s & 2s & 3s & Avg. & 1s & 2s & 3s & Avg. & 1s & 2s & 3s & Avg. \\ \hline
IL & LiDAR & - & - & - & - & - & - & - & - & 0.44 & 1.15 & 2.47 & 1.35 \\
NMP & LiDAR & - & - & - & - & - & - & - & - & 0.53 & 1.25 & 2.67 & 1.48 \\
FF & LiDAR & - & - & - & - & - & - & - & - & 0.55 & 1.20 & 2.78 & 1.43 \\ \hline
UniAD & Camera & - & - & - & - & - & - & - & - & 0.48 & 0.96 & 1.65 & 1.03 \\
VAD-Base & Camera & - & - & - & - & - & - & - & - & 0.54 & 1.15 & 1.98 & 1.22 \\
OccNet & Camera & - & - & - & - & - & - & - & - & 1.29 & 2.13 & 2.99 & 2.14 \\
OccWorld-S & Camera & 21.09 & 16.17 & 4.95 & 5.00 & 0.28 & 0.26 & 0.24 & 0.26 & 0.67 & 1.69 & 3.13 & 1.83 \\
BevFormer+OccWorld  & Camera & 23.28  & 17.71  & 14.06 & 18.35   &   5.04 & 3.34 & 1.24  & 3.20 & 0.43  & 0.87  & 1.31  & 0.87 \\
BevDet+OccWorld & Camera & 24.12 & 18.19 & 15.44  &  19.25    & 6.21  & 4.01  &1.39  & 3.87  & 0.41  & 0.84  & 1.28  & 0.84 \\
FBOCC+OccWorld & Camera & 24.22  & 18.49  & 15.64  & 19.45 & 6.55  & 4.24  &1.44  & 4.07 & 0.37  &  0.77 & 1.14  &  0.76\\
\textbf{BevFormer+Ours} & Camera & 25.35 & 21.09 & 16.17 & 20.87 & 9.11 & 7.98 & 5.02 & 7.37 & 0.26 & 0.67 & 0.98 & 0.64 \\
\textbf{BevDet+Ours} & Camera & 27.07 & 23.42 & 18.56 & 23.02 & 10.99 & 9.92 & 8.78 & 9.90 & 0.23 & 0.48 & 0.74 & 0.48 \\
\textbf{FBOCC+Ours} & Camera & \textbf{27.11} & \textbf{24.07} & \textbf{20.19} & \textbf{23.79} & \textbf{11.28} & \textbf{10.21} & \textbf{9.13} & \textbf{10.21} & \textbf{0.21} & \textbf{0.40} & \textbf{0.67} & \textbf{0.43} \\ \hline
\emph{Vanilla} & Occ. & 21.36 & 18.31 & 14.82 & 18.16 & 14.15 & 9.80 & 6.77 & 10.24 & 0.48 & 0.62 & 0.79 &  0.63  \\
OccNet & Occ. & - & - & - & - & - & - & - & - & \multicolumn{1}{l}{1.29} & \multicolumn{1}{l}{2.31} & \multicolumn{1}{l}{2.98} & \multicolumn{1}{l}{2.25} \\
OccWorld & Occ. & 34.63 & 25.07 & 20.18 & 26.63 & 25.78 & 15.14 & 10.51 & 17.14 & 0.43 & 1.08 & 1.99 & 1.17 \\
\textbf{Ours} & Occ. & \textbf{36.65} & \textbf{32.14} & \textbf{28.77} & \textbf{32.52} & \textbf{24.02} & \textbf{21.65} & \textbf{17.29} & \textbf{20.99} & \textbf{0.12} & \textbf{0.24} & \textbf{0.49} & \textbf{0.28} \\ \hline
\end{tabular}
}   
\end{table*}

\begin{figure*}[t]
    \centering
    \includegraphics[width=0.85\textwidth]{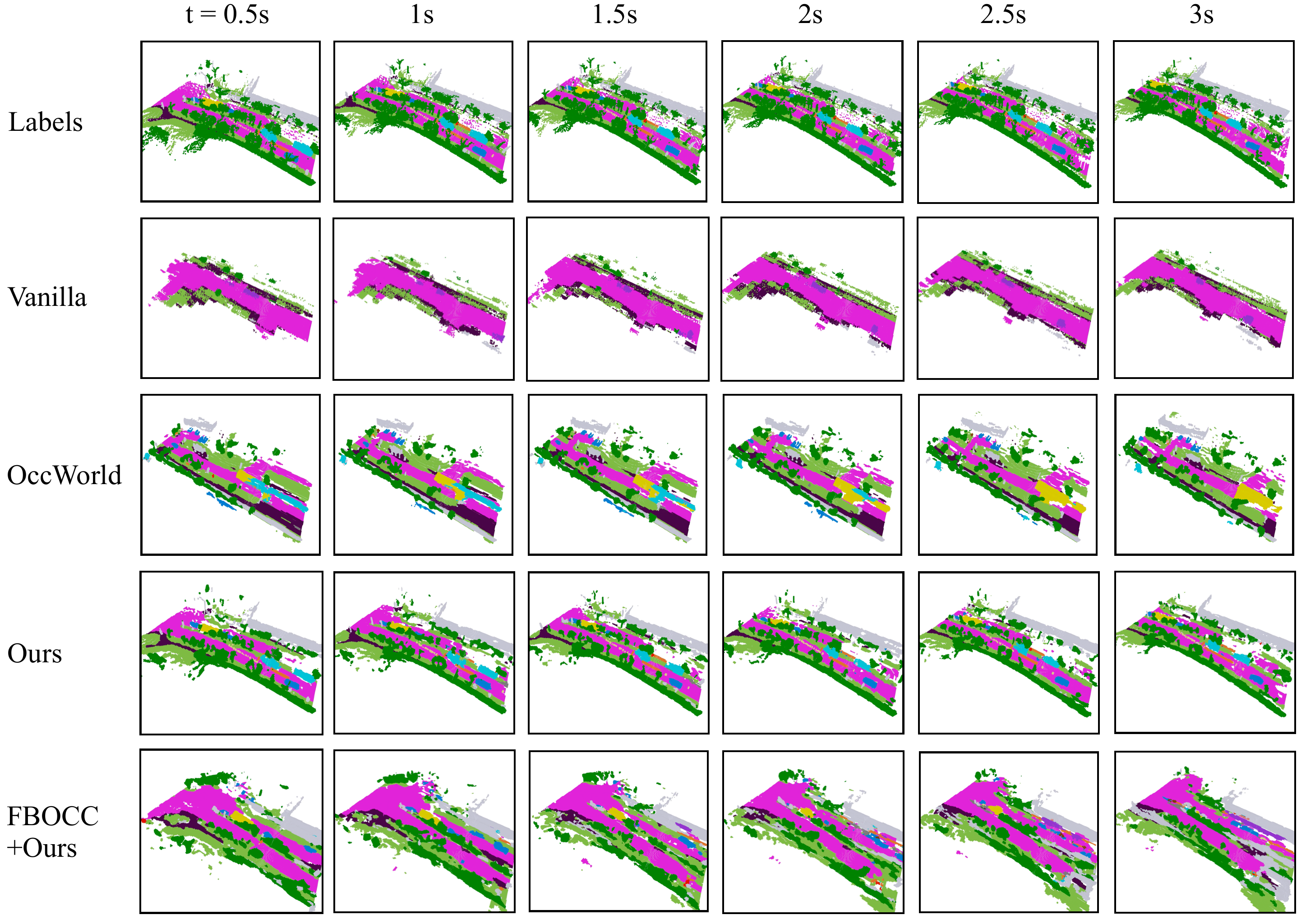}
    \caption{Qualitative 4-D occupancy forecasting results of our Occ-LLM. ``Vanilla" refers to the direct flattening of occupancy representation and its injection into the LLM for training \textbf{(zoom in for the best view)}.}
    \label{fig:forecast}
\end{figure*}

\subsubsection{4-D occupancy forecasting and self-ego planning}

Table~\ref{tab:quanti} compares our methods with state-of-the-art approaches in 4D occupancy forecasting and motion planning, providing metrics such as IoU, mIoU, and L2 distance at 1, 2, and 3-second intervals. Our methods consistently outperform the state-of-the-art in accuracy and consistency, as shown in Fig.~\ref{fig:forecast}.

The evaluated methods include LiDAR-based approaches like IL \cite{IL}, NMP \cite{NMP}, and FF \cite{FF}, as well as camera-based methods such as UniAD \cite{hu2023_uniad}, VAD-Base \cite{jiang2023vad}, and OccNet \cite{occnet}. We also integrate predicted occupancy data into our Occ-LLM framework, achieving higher performance with models like BevFormer+Ours, which reaches an average IoU of 23.79\%, mIoU of 10.21\%, and an L2 distance of 0.43 meters.

Compared to occupancy-based methods, our approach surpasses OccWorld, with an average IoU of 32.52\%, mIoU of 20.99\%, and an L2 distance of 0.28 meters, demonstrating superior accuracy and reliability for autonomous driving.

\begin{figure*}[t]
    \centering
    \includegraphics[width=0.85\textwidth]{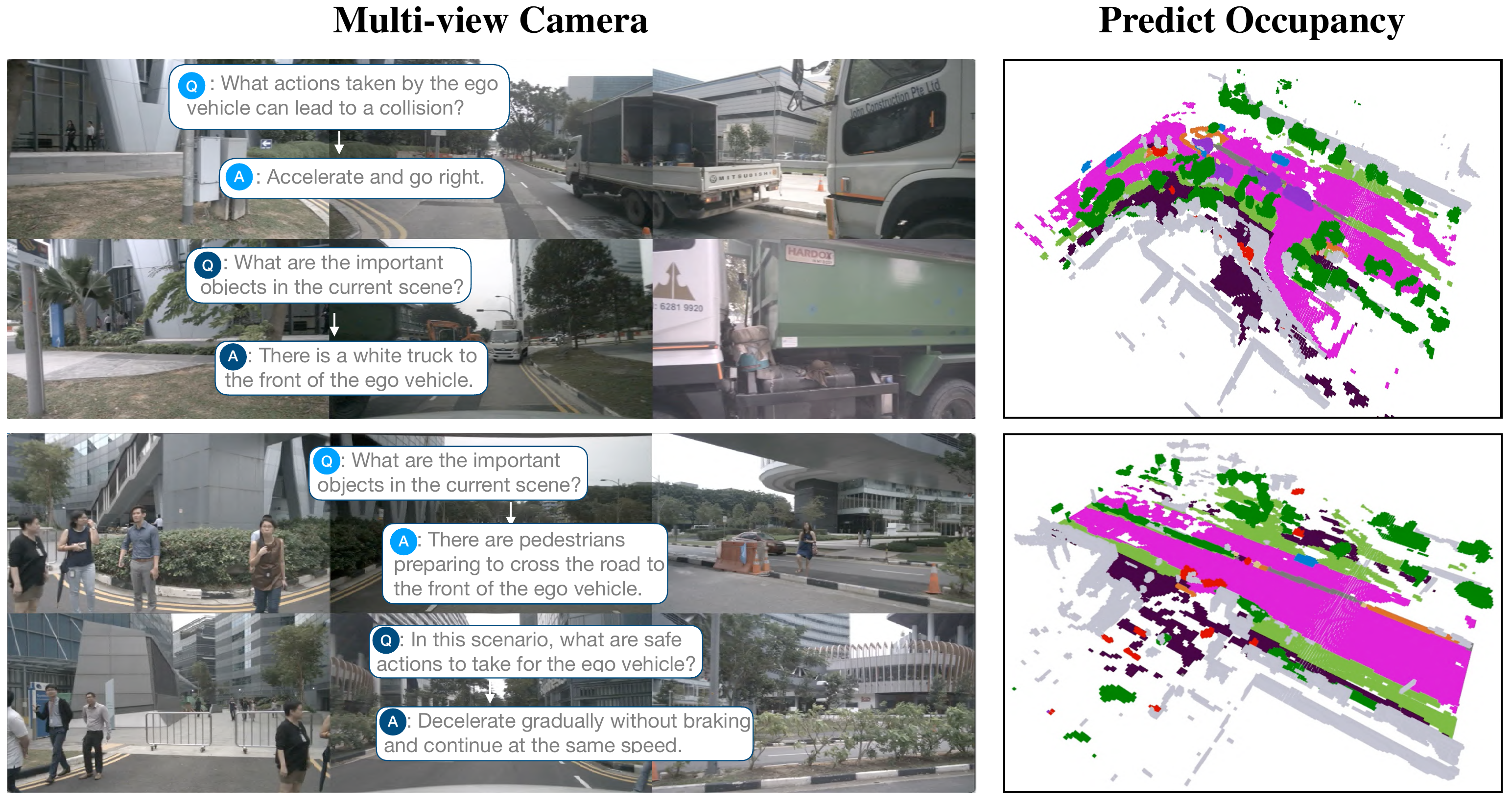}
    \caption{Qualitative question-answering results of our Occ-LLM. The left panel displays the raw scene data, while the right panel shows the predicted occupancy generated by FBOCC \cite{li2023fbocc}. The questions (Q) and the corresponding predicted answers (A) are illustrated \textbf{(zoom in for the best view)}.}
    \label{fig:qa}
    \vspace{-2mm}
\end{figure*}

\subsubsection{Question-answering}

Our proposed method demonstrates advanced question-answering capabilities specifically tailored for autonomous driving scenarios. As illustrated in Figure~\ref{fig:qa}, the system effectively interprets multi-view camera inputs to predict occupancy and provide accurate responses to queries regarding the driving environment. It can identify critical objects in the scene, recommend safe maneuvers for the ego vehicle, and describe potential hazards, such as pedestrians preparing to cross the road.

To quantitatively assess our system's performance, we conducted a comparative evaluation against the DriveLM model \cite{sima2023drivelm} using standard metrics, namely BLEU \cite{papineni2002bleu}, ROUGE\_L \cite{lin2004rouge}, CIDEr \cite{vedantam2015cider}, and GPT Score \cite{openai2023gpt4}, as presented in Table~\ref{tab:qa-quant}. Details of these evaluation metrics are provided in \cite{sima2023drivelm}. Our model outperforms DriveLM across all metrics, achieving superior scores. These results substantiate the effectiveness of our approach in delivering accurate and contextually relevant answers within autonomous driving environments.

% Our proposed method exhibits advanced question-answering capabilities tailored for autonomous driving scenarios. As illustrated in Figure~\ref{fig:qa}, the system effectively interprets multi-view camera inputs to predict occupancy and provide accurate responses to queries about the driving environment. It can identify critical objects in the scene, recommend safe maneuvers for the ego vehicle, and describe potential hazards such as pedestrians preparing to cross the road.

% To quantitatively evaluate our system's performance, we compared it with the DriveLM \cite{sima2023drivelm} model using standard metrics such as BLEU \cite{papineni2002bleu}, ROUGE\_L \cite{lin2004rouge}, CIDEr \cite{vedantam2015cider}, and GPT Score \cite{openai2023gpt4} (the detail of evaluating metrics can be viewed in DriveLM), as presented in Table~\ref{tab:qa-quant}. Our model outperforms DriveLM across all metrics, achieving higher scores. These results demonstrate the effectiveness of our approach in providing accurate and contextually relevant answers in autonomous driving environments.

% \begin{tabular}{lllll}
% Methods &  &  &  &  \\
% DriveLM &  &  &  &  \\
% Ours &  &  &  & 
% \end{tabular}

\begin{table}[thb]
\centering
\caption{Quantitative evaluation of question-answering performance metrics comparing our Occ-LLM with DriveLM.}
\label{tab:qa-quant}
\scalebox{0.95}{
\begin{tabular}{lcccc}
\hline
Methods & Bleu & ROUGE\_L & CIDEr & GPT Score$\uparrow$ \\ \hline 
DriveLM & 0.68 & 0.74 & 0.32 & 52.91 \\
Ours & \textbf{0.71} & \textbf{0.75} & \textbf{2.95} & \textbf{56.72} \\ \hline
\end{tabular}}
\end{table}
% Our proposed methods demonstrate significant capabilities in question-answering within the context of autonomous driving. Figure~\ref{fig:qa} provides a visual demonstration of this capability. The system is designed to interpret multi-view camera inputs, predict occupancy, and respond accurately to various queries related to the driving environment. For instance, the system can identify critical objects in the scene, suggest safe actions for the ego vehicle, and describe potential hazards such as pedestrians preparing to cross the road.

\subsection{Ablation Study}\label{sec:ablation}
\subsubsection{Comparative analysis of OccWorld's VAE and the proposed MS-VAE}
Table~\ref{tab:ms-vae} compares OccWorld's VAE \cite{zheng2023occworld} with our proposed MS-VAE, showing significant improvements in reconstruction performance. The addition of 3D convolution layers and the Motion Separation strategy has increased IoU and mIoU, with MS-VAE achieving 62.74\% IoU and 71.08\% mIoU, compared to 59.07\% and 60.50\% for OccWorld's VAE.

\begin{table}[t]
\centering
\caption{Comparative analysis of OccWorld's VAE \cite{zheng2023occworld} and the proposed MS-VAE. The baseline model employs the same architecture as OccWorld's VAE. %Integrating our proposed modules has resulted in a substantial improvement in reconstruction performance. 
}
\label{tab:ms-vae}
\scalebox{0.93}{
\begin{tabular}{lcccc}
\hline
\multicolumn{1}{c}{\multirow{2}{*}{Methods}} & \multirow{2}{*}{Latent Shape}  & \multirow{2}{*}{Parameters(M)} & \multicolumn{2}{c}{Reconstruction} \\ \cline{4-5} 
\multicolumn{1}{c}{}                         &                             &                              & IoU $\uparrow$              & mIoU $\uparrow$           \\ \hline
OccWorld        & 50,50,128                     & 14.28                               & 59.07 & 60.50	            \\ \hline
Baseline                                 & 50,50,32                  & 2.25                              &  57.90   & 59.34                 \\
+3D Conv.                                & 50,50,32                     & 2.30                            &    61.94     & 65.81 \\
+Motion Separation                & 50,50,64                    & 2.30                            & \textbf{62.74}            & \textbf{71.08}           \\ \hline
\end{tabular}}
\end{table}

\begin{table}[thb]
\centering
\caption{Comparative analysis of different patch sizes in patchify. Each value represents the average over a 3-second interval.}
\label{tab:patchsize}
\scalebox{0.93}{
\begin{tabular}{lcccc}
\hline
\multirow{2}{*}{Patch size} & \multicolumn{2}{c}{Trainset} & \multicolumn{2}{c}{Valset} \\ \cline{2-5} 
 & IoU $\uparrow$ & mIoU $\uparrow$ & IoU $\uparrow$ & mIoU $\uparrow$ \\ \hline
1 & 20.91 & 15.14 & 16.46 & 7.71 \\
5 & 28.94 & 22.61 & 26.55 & 25.81 \\
10 & \textbf{32.48} & \textbf{26.16} & \textbf{27.12} & \textbf{26.83} \\
25 & 25.97 & 19.69 & 16.33 & 11.89 \\ \hline
\end{tabular}}
\end{table}

\begin{table}[th]
\centering
\caption{Ablation study of Occ-LLM modules. Each value represents the performance on the validation set.}
\label{tab:ablation-occ-llm}
\scalebox{0.95}{
\begin{tabular}{lccc}
\hline
\multirow{2}{*}{Modules} & \multicolumn{3}{c}{Valset} \\ \cline{2-4} 
 & IoU $\uparrow$ (\%) & mIoU $\uparrow$ (\%) & L2 $\downarrow$ (m) \\ \hline 
Baseline & 20.67 & 16.63 & 0.82 \\
+Pre Fusion & 24.44 & 18.27 & 0.69 \\
+Motion Separation & \textbf{32.52} & \textbf{20.99} & \textbf{0.28} \\ \hline
\end{tabular}}
\end{table}

\subsubsection{Comparative analysis of different patch sizes in patchify}
Table~\ref{tab:patchsize} examines the effect of varying patch sizes on reconstruction performance. A patch size of 10 performs best, with an IoU of 32.48\% and mIoU of 26.16\% on the Trainset, and 27.12\% and 26.83\% on the Valset, balancing detail capture and efficiency.

\subsubsection{Ablation study of Occ-LLM modules}
Table~\ref{tab:ablation-occ-llm} shows an ablation study of Occ-LLM modules. The baseline achieves 20.67\% IoU, 16.63\% mIoU, and 0.82m L2 distance. Adding the Pre Fusion module improves these metrics, and incorporating the Motion Separation (MS) module further boosts IoU to 32.52\%, mIoU to 20.99\%, and reduces L2 distance to 0.28m, highlighting the benefits of the MS module.

\section{CONCLUSION}

This paper introduces the Occupancy-based Large Language Model (Occ-LLM), which enhances autonomous driving by integrating LLMs with occupancy representation. It proposes the Motion Separation Variational Autoencoder (MS-VAE) to address category imbalance by separating dynamic objects from static scenes. Occ-LLM outperforms state-of-the-art methods in 4D occupancy forecasting, self-ego planning, and scene question answering, achieving higher Intersection over Union (IoU) and mean IoU (mIoU) scores and reducing planning errors.

%%%%%%%%%%%%%%%%%%%%%%%%%%%%%%%%%%%%%%%%%%%%%%%%%%%%%%%%%%%%%%%%%%%%%%%%%%%%%%%%

%%%%%%%%%%%%%%%%%%%%%%%%%%%%%%%%%%%%%%%%%%%%%%%%%%%%%%%%%%%%%%%%%%%%%%%%%%%%%%%%

%%%%%%%%%%%%%%%%%%%%%%%%%%%%%%%%%%%%%%%%%%%%%%%%%%%%%%%%%%%%%%%%%%%%%%%%%%%%%%%%
% \section*{APPENDIX}

% Appendixes should appear before the acknowledgment.

% \section*{ACKNOWLEDGMENT}

% The preferred spelling of the word ÒacknowledgmentÓ in America is without an ÒeÓ after the ÒgÓ. Avoid the stilted expression, ÒOne of us (R. B. G.) thanks . . .Ó  Instead, try ÒR. B. G. thanksÓ. Put sponsor acknowledgments in the unnumbered footnote on the first page.

% %%%%%%%%%%%%%%%%%%%%%%%%%%%%%%%%%%%%%%%%%%%%%%%%%%%%%%%%%%%%%%%%%%%%%%%%%%%%%%%%

% References are important to the reader; therefore, each citation must be complete and correct. If at all possible, references should be commonly available publications.

\bibliographystyle{abbrv}
\bibliography{ref}

\end{document}